\def\year{2020}\relax
\documentclass[letterpaper]{article} 
\usepackage{aaai20}  
\usepackage{times}  
\usepackage{helvet} 
\usepackage{courier}  
\usepackage[hyphens]{url}  
\usepackage{graphicx}  
\usepackage{latexsym}
\usepackage[utf8]{inputenc}
\usepackage{amsmath}
\DeclareMathOperator*{\argmin}{arg\,min}
\usepackage{booktabs}
\usepackage{paralist}
\usepackage[inline]{enumitem}
\usepackage[export]{adjustbox}

\usepackage[algoruled,linesnumbered,vlined]{algorithm2e}

\usepackage{array}
\usepackage{multirow} 
\usepackage{cases}
\usepackage{amsfonts}
\usepackage{soul}
\usepackage{colortbl}
\usepackage{xspace}
\definecolor{shadecolor}{gray}{0.85}

\def \our {\mbox{HAMNER}\xspace}
\def \ncbi {\mbox{\textsf{NCBI-Disease}}\xspace}
\def \bccdr {\mbox{\textsf{BC5CDR}}\xspace}
\def \laptop {\mbox{\textsf{LaptopReview}}\xspace}

\def \bccdrthehp {\mbox{85.46}\xspace}
\def \bccdrthehr {\mbox{86.78}\xspace}
\def \bccdrthehf {\mbox{86.12}\xspace}

\def \ncbithehp {\mbox{80.63}\xspace}
\def \ncbithehr {\mbox{83.94}\xspace}
\def \ncbithehf {\mbox{82.24}\xspace}

\def \laptopthehp {\mbox{74.02}\xspace}
\def \laptopthehr {\mbox{62.02}\xspace}
\def \laptopthehf {\mbox{67.46}\xspace}

\def \bccdrthep {\mbox{86.01}\xspace}
\def \bccdrther {\mbox{86.34}\xspace}
\def \bccdrthef {\mbox{86.17}\xspace}

\def \ncbithep {\mbox{82.03}\xspace}
\def \ncbither {\mbox{83.56}\xspace}
\def \ncbithef {\mbox{82.79}\xspace}

\def \laptopthep {\mbox{74.02}\xspace}
\def \laptopther {\mbox{62.02}\xspace}
\def \laptopthef {\mbox{67.46}\xspace}
\def \bccdrtep {\mbox{85.98}\xspace}
\def \bccdrter {\mbox{85.95}\xspace}
\def \bccdrtef {\mbox{85.97}\xspace}

\def \ncbitep {\mbox{79.91}\xspace}
\def \ncbiter {\mbox{80.25}\xspace}
\def \ncbitef {\mbox{80.08}\xspace}

\def \laptoptep {\mbox{71.78}\xspace}
\def \laptopter {\mbox{59.82}\xspace}
\def \laptoptef {\mbox{65.22}\xspace}
\def \bccdrthp {\mbox{83.04}\xspace}
\def \bccdrthr {\mbox{85.07}\xspace}
\def \bccdrthf {\mbox{84.04}\xspace}

\def \ncbithp {\mbox{80.13}\xspace}
\def \ncbithr {\mbox{81.81}\xspace}
\def \ncbithf {\mbox{80.95}\xspace}

\def \laptopthp {\mbox{69.39}\xspace}
\def \laptopthr {\mbox{59.42}\xspace}
\def \laptopthf {\mbox{64.00}\xspace}
\def \bccdrtp {\mbox{83.28}\xspace}
\def \bccdrtr {\mbox{83.80}\xspace}
\def \bccdrtf {\mbox{83.54}\xspace}

\def \ncbitp {\mbox{79.55}\xspace}
\def \ncbitr {\mbox{76.56}\xspace}
\def \ncbitf {\mbox{78.02}\xspace}

\def \laptoptp {\mbox{69.74}\xspace}
\def \laptoptr {\mbox{57.16}\xspace}
\def \laptoptf {\mbox{62.78}\xspace}

\title{\our: Headword Amplified Multi-span Distantly Supervised Method for
  Domain Specific Named Entity Recognition%
  \thanks{S.~Liu and W.~Wang are the co-corresponding authors.}
}
\author{Shifeng Liu\textsuperscript{\rm 1} 
  Yifang Sun\textsuperscript{\rm 1} 
  Bing Li\textsuperscript{\rm 1} 
  Wei Wang\textsuperscript{\rm 1,2} 
  Xiang Zhao\textsuperscript{\rm 3}\\
  \textsuperscript{\rm 1}University of New South Wales,
  \textsuperscript{\rm 2}Dongguan University of Technology\\
  \textsuperscript{\rm 3}National University of Defence Technology\\
  \{shifeng.liu, bing.li\}@unsw.edu.au,
  \{yifangs, weiw\}@cse.unsw.edu.au,
  xiangzhao@nudt.edu.cn
}

\begin{document}

\maketitle

\begin{abstract}
To tackle Named Entity Recognition~(NER) tasks, supervised methods need to obtain 
sufficient cleanly annotated data, which is labor and time consuming. 
On the contrary, distantly supervised methods acquire automatically annotated data
using dictionaries to alleviate this requirement.
Unfortunately, dictionaries 
hinder the effectiveness of distantly supervised methods for NER due to its limited 
coverage, especially in specific domains.
In this paper, we aim at the limitations of the dictionary usage and mention boundary detection.
We generalize the distant supervision by extending the dictionary with headword based non-exact matching.
We apply a function to better weight the matched entity mentions.
We propose a span-level model, which classifies all the possible spans then infers the
selected spans with a proposed dynamic programming algorithm.
Experiments on all three benchmark datasets demonstrate that our method outperforms 
previous state-of-the-art distantly supervised methods.
\end{abstract}

\section{Introduction}
\label{sec:introduction}
Named entity recognition~(NER) is a task that extracts entity mentions from sentences and
classifies them into pre-defined types, such as person, location, disease,
chemical, etc. It is a vital task in natural language processing~(NLP), which
benefits many downstream applications including relation
extraction~\cite{DBLP:conf/acl/MintzBSJ09}, event
extraction~\cite{DBLP:conf/aaai/NguyenG18}, and
co-reference resolution~\cite{DBLP:conf/emnlp/ChangSR13}.  


With a sufficient amount of cleanly annotated texts (i.e., the training data),
supervised
methods~\cite{DBLP:conf/naacl/LampleBSKD16,DBLP:conf/acl/MaH16}
have shown their ability to achieve high-quality performance in general domain
NER tasks and benchmarks. However, obtaining
cleanly annotated texts is labor-intensive and time-consuming, especially in
specific domains, such as the biomedical domain and the technical domain. This
hinders the usage of supervised methods in real-world applications.

Distantly supervised methods circumvent the above issue by generating pseudo
annotations according to \emph{domain specific dictionaries}. Dictionary is a
collection of $\langle{\text{\textit{entity mention}, \textit{entity type}}}\rangle$
pairs. Distantly supervised methods firstly find entity mentions by exact string
matching~\cite{DBLP:conf/wassa/GiannakopoulosM17,DBLP:conf/emnlp/ShangLGRR018}
or regular expression matching~\cite{DBLP:journals/corr/Fries0RR17} with the
dictionary, and then assign corresponding types to the mentions.
A model is trained on the training corpus
with the pseudo annotations. As the result, distant supervision significantly
reduces the annotation cost, while not surprisingly, the accuracy (e.g., 
precision and recall) reduces.

In this paper, we aim to reduce the gap between distantly supervised methods
and supervised methods.
We observed two limitations in distantly supervised methods.
The first limitation is that the information in the dictionary has not been fully mined and used.
For example, consider a newly discovered disease namely \textit{ROSAH syndrome},
 which is unlikely to exist in the
dictionaries, and hence cannot be correctly  
extracted and annotated if we use simple surface matching. 
However, human beings can easily recognize it as a disease, since there are many
disease entity mentions in the dictionaries that are ended with \textit{syndrome}. 
This motivates us to use \emph{headwords} of entity mentions as indicators of
entity types, and thus improves the quality of the pseudo annotations.

The second limitation in distantly supervised methods is that 
most of the errors come
from incorrect boundaries\footnote{For example, even the state-of-the-art
  distantly supervised method~\cite{DBLP:conf/emnlp/ShangLGRR018} has at least
  40\% errors coming from incorrect boundaries, on all three benchmarks that are
  evaluated in this paper.}.
Most methods (including supervised methods) model the NER problem as a sequence
labeling task and use popular architectures/classifiers like
BiLSTM-CRF~\cite{DBLP:conf/acl/MaH16}. However, CRFs suffer from sparse
boundary tags~\cite{DBLP:conf/ijcai/0034YS19}, and pseudo annotations can only
be more sparse and noisy. In addition, CRFs focus more on word-level
information and thus cannot make full use of span-level
information~\cite{DBLP:conf/acl/ZhuoCZZN16}.
Some methods choose to fix the entity boundaries before predict the entity type.
Apparently, any incorrect boundary will definitely lead to incorrect output, no
matter how accurate the subsequent classifier is.
Therefore, we propose to decide entity boundaries 
after predicting entity types. As such, there would be more information, 
such as the types and confidence scores of entity mentions, which
can help to determine more accurate entity boundaries.

Based on the above two ideas, we propose a new distantly supervised method
named \our (Headword Amplified Multi-span NER) for NER tasks in specific
domains. 
We first introduce a novel dictionary extension method based on the semantic
matching of headwords.
To account for the possible noise introduced by the extended entity mentions,
we also use the similarity between the headwords of 
the extended entity mentions and the original entity mentions to represent the 
quality of the extended entity mentions. 
The extended dictionary will be used to generate pseudo annotations. 
We then train a model to estimate the type of a given span from a sentence
based on its contextual information.
Given a sentence, \our uses the trained model to predict types of \emph{all}
the possible spans subject to the pre-defined maximum number of words, 
and uses a dynamic programming based inference algorithm
to select the most proper boundaries and types of entity mentions while suppressing
overlapping and spurious entity mentions.

The main contributions of this paper are 
\begin{itemize}[noitemsep,leftmargin=*]
\item We generalize the distant supervision idea for NER by extending the
  dictionaries using semantic matching of headwords. Our extension is
  grounded in linguistic and distributional semantics. We use the extended
  dictionary to improve the quality of the pseudo annotations. 
  \item We propose a span-level model with both span information and contextual 
  information to predict the type for a given span. We propose a dynamic 
  programming inference algorithm to select the spans which are the most
  likely to be entity mentions.

  \item Experiments on three benchmark datasets have demonstrated that \our
    achieves the best performance with dictionaries only and no human efforts.
    Detailed analysis has shown the effectiveness of our designed method.

\end{itemize}

\section{Related Work}
\label{sec:related work}
Named entity recognition~(NER) attracts researchers and has been tackled by both
supervised and semi-supervised methods. 
Supervised methods, including feature based
methods~\cite{DBLP:conf/conll/RatinovR09,DBLP:conf/pakdd/LiuSWZ18,DBLP:conf/dasfaa/SunLWW19} and neural
network based
methods~\cite{DBLP:conf/naacl/LampleBSKD16,DBLP:conf/acl/MaH16,DBLP:journals/corr/abs-1812-09449}, 
require cleanly annotated texts. 
Semi-supervised methods either utilize more unlabeled
data~\cite{DBLP:conf/naacl/PetersNIGCLZ18} or generate annotated
data gradually~\cite{DBLP:conf/acl/TomanekH09,DBLP:conf/ijcnn/HanKK16,DBLP:conf/acl/BrookeHB16}.  

Distant supervision is proposed to alleviate human efforts in data annotation~\cite{DBLP:conf/acl/MintzBSJ09,DBLP:journals/jmlr/WallaceKSZM16}.
\citeauthor{DBLP:conf/acl/MintzBSJ09}~\citeyear{DBLP:conf/acl/MintzBSJ09}~propose distant supervision to handle the limitation of human annotated data for relation extraction task~\cite{DBLP:conf/acl/MintzBSJ09}.
They use Freebase~\cite{DBLP:conf/sigmod/BollackerEPST08}, instead of human, to generate training data with heuristic matching rules.
For each pair of entity mentions with some Freebase relations, sentences with these entity mentions are regarded as sentences with such relations.
Beyond the heuristic matching rules, \citeauthor{DBLP:journals/jmlr/WallaceKSZM16}~\citeyear{DBLP:journals/jmlr/WallaceKSZM16}~learn a matching function with a small amount of manually annotated data~\cite{DBLP:journals/jmlr/WallaceKSZM16}.

Recently, distant supervision has been explored for NER tasks~\cite{DBLP:journals/corr/Fries0RR17,DBLP:conf/wassa/GiannakopoulosM17,DBLP:conf/emnlp/ShangLGRR018}.
SwellShark~\cite{DBLP:journals/corr/Fries0RR17} utilizes a collection of
dictionaries, ontologies, and, optionally, heuristic rules to generate annotations
and predict entity mentions in the biomedical domain without human annotated data.
\citeauthor{DBLP:conf/emnlp/ShangLGRR018}~\citeyear{DBLP:conf/emnlp/ShangLGRR018}~use exact string matching to generate pseudo annotated data, 
and apply high-quality phrases~(i.e., mined in the same domain without assigning
any entity types) to reduce the number for false negative
annotations~\cite{DBLP:conf/emnlp/ShangLGRR018}. 
However, their annotation quality is limited by the coverage of the dictionary,
which leads to a relatively low recall. 
In the technical domain, Distant-LSTM-CRF~\cite{DBLP:conf/wassa/GiannakopoulosM17}
applies syntactic rules and pruned high-quality phrases to generate
pseudo annotated data for distant supervision. 
The major differences between \our and other distantly supervised methods are
\begin{enumerate*}
  \item \our extended the coverage of dictionaries without human efforts.
  \item \our makes predictions in the span level with both entity type and boundary information.
\end{enumerate*}

\section{Problem Definition}
\label{sec:problem}
We represent a sentence as a word sequence $(x_1, x_2, \ldots, x_N)$.
For span $(x_i, \ldots, x_j)$ from the sentence, we use $\langle i, j\rangle$
to denote its boundaries, and use $l\in L$ to denote its type,
where $L$ represents the list of \emph{pre-defined} types (e.g.,
\texttt{Disease}, \texttt{Chemical}) and \emph{none} type (e.g., \texttt{None}). 
We let \texttt{None} be the last element in $L$ (i.e., $L_{|L|}$).

We tackle the problem with distant supervision.
Unlike supervised and semi-supervised methods, we require no human annotated
training data. Instead, we only require a dictionary as the input in addition to
the raw text. The dictionary is a collection of $\langle$entity mention, entity
type$\rangle$-tuples. We use dictionary in the training phase to help generate
pseudo annotations on the training corpus.

We argue that dictionaries are easy to obtain, either from publicly available
resources, such as Freebase~\cite{DBLP:conf/sigmod/BollackerEPST08} and SNOMED
Clinical Terms\footnote{\url{http://www.snomed.org}}, or by crawling terms from
some domain-specific high-quality websites~\footnote{e.g.,
  \url{https://www.computerhope.com/jargon.htm} for computer terms, and
  \url{https://www.medilexicon.com/abbreviations} for medical terms}.

\section{The Proposed Method}
\label{sec:method}
\begin{figure}[ht]
    \centering
    \includegraphics[width=0.8\columnwidth]{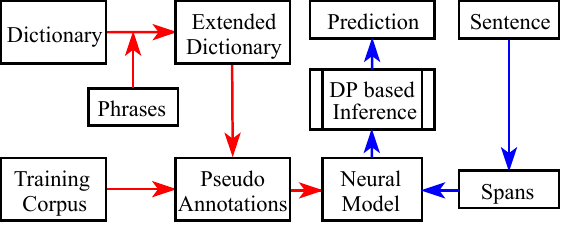}
    \caption{The process of the proposed method. Red arrows show the training steps, including dictionary extension, pseudo annotation generation, and model training. Blue arrows show the prediction phases, including span generation, model prediction, and dynamic programming~(DP) based inference.}
    \label{fig:anno}
\end{figure}
Figure~\ref{fig:anno} illustrates the key steps in HAMNER. 
In training, our method firstly generates pseudo annotations according to a
headword-based extended dictionary (details in Section~\ref{sec:overview}). Then
a neural model is trained using the pseudo annotated data. The model takes a span
and its context in the sentence as input, and predicts the type of the span.
The structure of the neural model is introduced in
Section~\ref{sec:modell-type-distr}.

Prediction is performed in two phases. Given a sentence, in phase one, we generate all
the possible spans whose length is no more than a specified threshold, and
use the trained neural model to predict the types of these spans.
In phase two, we apply a dynamic programming based inference algorithm to
determine the entity mentions and their types. The details are
presented in Section~\ref{sec:inference}.

Unlike previous
works~\cite{DBLP:conf/naacl/PetersNIGCLZ18,DBLP:conf/wassa/GiannakopoulosM17}, 
we solve NER in span-level where information has not been fully explored. 
HAMNER is based on all the possible spans up to a certain length instead of
only considering those based on the prediction of a
model~\cite{DBLP:conf/emnlp/ShangLGRR018}. 
While this is an effective solution for purely supervised methods in 
semantic role labeling~(SRL)~\cite{DBLP:conf/emnlp/OuchiS018} and 
nested/non-overlap NER~\cite{DBLP:conf/emnlp/SohrabM18,DBLP:conf/acl/XiaZYLDWFMY19},
we are the first to apply this to distantly supervised NER.
Specifically, we generate all the spans $\langle i,j\rangle$ containing up to
$M$ words as candidate spans, where $M$ is a pre-defined parameter. Therefore,
a sentence with $N$ words will generate $\frac{M(2N-M+1)}{2}$ spans, which is
still linear in the length of the sentence.


\subsection{Generating Pseudo Annotations}
\label{sec:overview}
In this section, we aim to improve the quality of the pseudo annotations.
While most of the existing distantly supervised methods use domain specific
dictionaries to annotate texts, we find that they are struggling because of the following two
cases.
\begin{itemize}
\item \textbf{Out-of-dictionary entity mentions}. It is common that the dictionaries are not frequently updated.
  However, new entities and concepts are generated everyday. Therefore, the coverage of a dictionary is generally not high.
\item \textbf{Synonyms and spelling differences}. Most dictionaries may not
  contain both terms in a pair of synonyms. 
  And they usually stick with one spelling form (e.g., British or American
  spelling). Both of these cases can enrich the annotations thus should not be
  ignored. 
\end{itemize}

Therefore, we propose to extend the dictionary by adding a set of domain
specific high-quality phrases. The high-quality phrases can be obtained in
multiple ways. For example, we can use an unsupervised phrase mining
methods~\cite{DBLP:conf/emnlp/ShangLGRR018,DBLP:conf/aaai/LiYWC17,DBLP:conf/ijcai/00020WWCZ18,DBLP:journals/tkde/LiYZWLZ19} on a large corpus in the same
domain, or collect unclassified terms from lexicon resources, such as
mediLexicon~\footnote{\url{www.medilexicon.com}}.

However, these phrases do not have types.
To assign types to these high-quality phrases, we propose to make use of headwords.
It is well-known that
headwords of entity mentions play an important role in information extraction 
tasks~\cite{DBLP:conf/acl/SurdeanuHWA03}.
In a similar vein, the headword of a noun phrase is usually highly indicative of
the type of noun phrase. For example, \textit{appendix cancer} shares the same
headword \textit{cancer} with \textit{liver cancer}; hence, if
we know that the type of the former is \texttt{Disease}, then we can infer that the
type of the latter is probably the same.

We also consider non-exact string matching on the headwords to deal with synonyms and spelling
differences. 
Our idea is based on the hypothesis that similar words tend to occur
in similar contexts~\cite{harris1954distributional} and hence may have a high probability of belonging to the
same type. Thanks to word embedding methods, we can measure such distributional
similarity based on the cosine similarity between the corresponding word
embeddings. For example, since the word embeddings of \textit{tumor} and
\textit{tumour} are close, their types should be the same.

\begin{algorithm}[htbp]
\SetKwFor{ForEach}{for each}{do}{end for}%
\SetKwInOut{Input}{Input}
\SetKwInOut{Output}{Output}

\small%
\Input{Dictionary $D$, High-quality Phrase Set $P$, headword frequency threshold
$\tau_1$, headword similarity threshold $\tau_2$} 
\Output{Extended Dictionary $D_{ext}$}
$H \gets$ set of headwords that occurs more than $\tau_1$ times in D;\label{alg:1}

$D_{ext} \gets \emptyset$;

\ForEach{$p \in P$}{
  \tcc{$s$: the max cosine similarity}
  \tcc{$T$: the set of types}
  $s \gets 0$;

  $T \gets \emptyset$;

  \ForEach{$\langle e_i, t_i\rangle \in D$}{
    \If{$hw(e_i) \in H \wedge sim(hw(p),hw(e_i)) \geq \tau_2$}{\label{alg:2}
      \If{$sim(hw(p),hw(e_i)) > s$}{\label{alg:3}
        $s \gets sim(hw(p),hw(e_i))$;

        $T \gets \{t_i\}$;
      }
      \ElseIf{$sim(hw(p),hw(e_i)) = s$}{
        $T \gets T \cup \{t_i\}$;\label{alg:4}
      }
    }
  }
  \ForEach{$t \in T$}{
    $D_{ext} = D_{ext} \cup \{(p, t, s)\}$;
  }
}
\ForEach{$\langle e_i, t_i\rangle \in D$}{
  $D_{ext} = D_{ext} \cup \{(e_i, t_i, 1)\}$;
}
\Return{$D_{ext}$}
  \caption{Dictionary Extension}
  \label{alg:dict-ext}
\end{algorithm}
Algorithm~\ref{alg:dict-ext} presents the procedure of dictionary extension in
HAMNER, where $hw(x)$ is the headword of the phrase/entity $x$, and $sim(x,y)$ is
the cosine similarity between the word embeddings of $x$ and $y$. We have
noticed that 
while the word embedding based non-exact string matching improves the 
coverage of the dictionary, it also brings some noise. Therefore,
we use $\tau_1$ to prune those infrequent headwords (i.e., Line~\ref{alg:1}),
and use $\tau_2$ to avoid matching with dissimilar headwords (i.e.,
Line~\ref{alg:2}). We also only keep the types with the highest cosine 
similarity (i.e., Line~\ref{alg:3}-\ref{alg:4}).

Through Algorithm~\ref{alg:dict-ext} we obtain a large set of
entity mentions, along with their types and the cosine similarities. We then
annotate the unlabeled data via exact string matching of the entity mentions.
We follow the existing work that favors longest matching entity mentions, hence we sort the
entity mentions in the dictionary by their length, and start matching the
spans in the sentences greedily from the longest one. We do not allow the
annotations to be nested or overlapping. But we do allow one span to have
multiple types as two entities of different types may share the same
entity mention in the dictionary.

In addition to the types, we also assign a weight to each annotation.
Assume that 
a span is matched by an entity mention in the extended dictionary 
with \emph{pre-defined} type $l$ and the corresponding
cosine similarity $s$. 
While we set annotation weight 1 for entity mentions themselves or their headwords appearing in the original dictionary,
we use a proposed $\mathrm{sigmoid}$ function to control the noise introduced by non-exact matched entity mentions.
Thus, we define the annotation weight $w_l$ for type $l$ using the function as follows, 
\begin{equation}
w_l = \begin{cases}
\mathrm{sigmoid}(\theta_1 \cdot s+ \theta_2), &\text{if s $<$ 1}.\\
1, &\text{if s $=$ 1}.
\end{cases}
\label{func:weight}
\end{equation}, where $\theta_1$ and $\theta_2$ are hyper-parameters.
The annotation weight can be interpreted as a confidence score of the annotation and will be
used in the neural model in Section~\ref{sec:modell-type-distr}.

For each span, we use $\langle i, j\rangle$ to indicate its boundaries
in the sentence, and hence it will be represented as $(\langle i, j\rangle,
w_{L_1},\dots, w_{L_{|L|}})$. 
If a span is annotated with pre-defined types, then we use
Equation~\eqref{func:weight} to compute the weights for the corresponding
annotated types, while setting weights to 0 for the rest types. 
Otherwise, only $w_{L_{|L|}}$~(i.e. weight of \texttt{None} type) is set to 1
while the rest weights set to 0. 
These two types of spans will serve as positive and negative samples during
training.

\subsection{Modelling Type Distribution of Spans}
\label{sec:modell-type-distr}

\begin{figure}[htbp]
  \includegraphics[width=0.9\columnwidth, left]{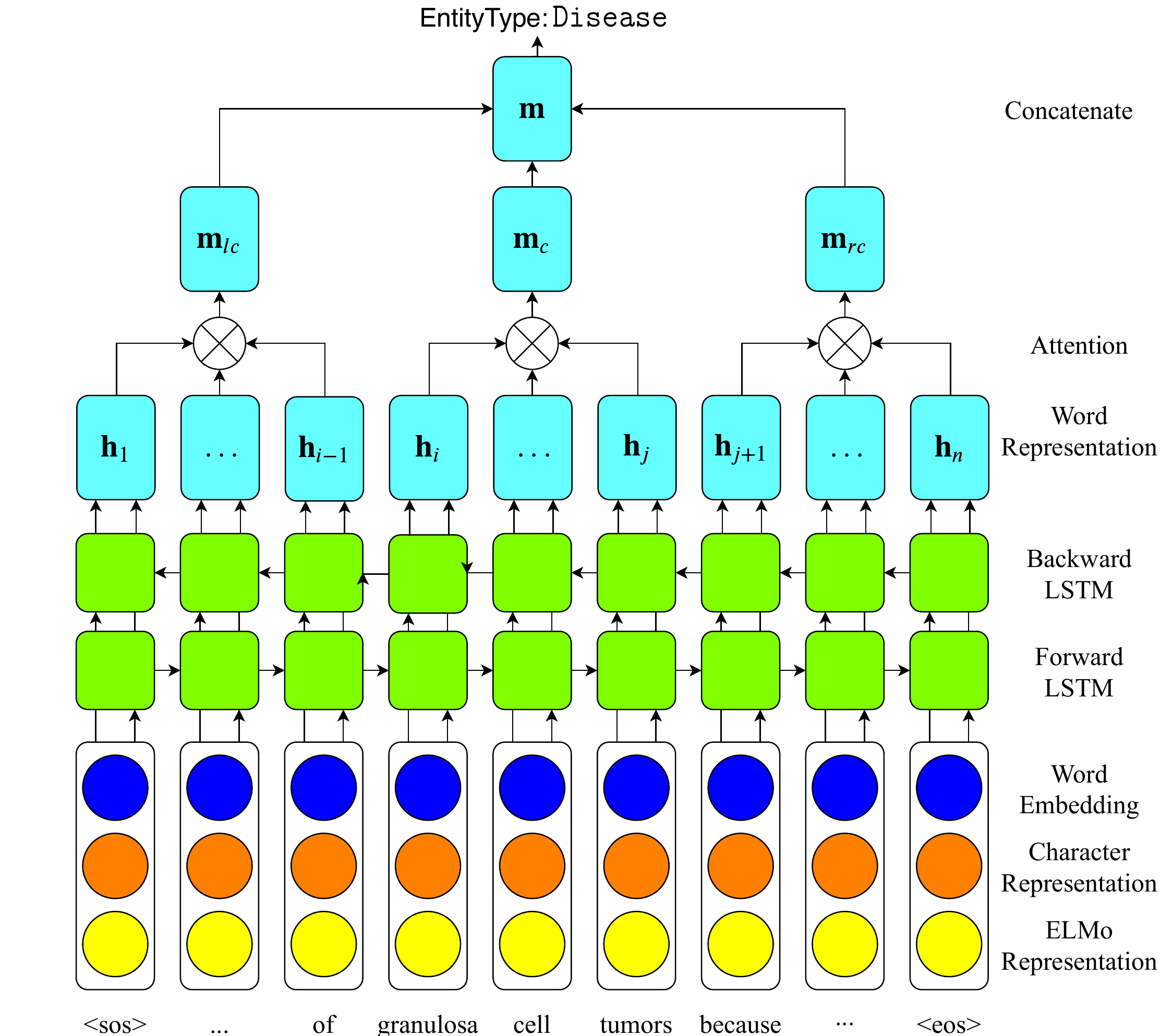}
  \caption{Overview of the neural model. The example shows the span (i.e.,
    \textit{granulosa cell tumors}) and the corresponding context. This span is
    annotated as \texttt{Disease} during preprocessing mentioned in
    Section~\ref{sec:overview}. Words $\langle sos\rangle$~(start of a sentence) and $\langle eos\rangle$~(end of a sentence) 
    are added in each sentence to construct contexts for spans at the start and at the end of a sentence.}
  \label{fig:framework}
\end{figure}
We design a novel neural network model to model the type distribution of given spans,
i.e., given any span $\langle{i, j}\rangle$, the model will predict
$p(l \mid \langle{i, j}\rangle$) for all types $l \in L$. %
Our model exploits typographical, semantics, and contextual features, and
employs a new loss function to deal with multiple weighted annotated types.
The architecture of our proposed neural model is shown in Figure~\ref{fig:framework}.

\subsubsection{Word Representation}

To extract morphological information, such as prefixes and suffixes of a word,  
we use \textbf{CNN} to encode the character representation
following~\cite{DBLP:conf/acl/MaH16}.
We also apply pre-trained \textbf{ELMo}~\cite{DBLP:conf/naacl/PetersNIGCLZ18}
for the contextualized word representation.

We then concatenate the pre-trained word embedding, ELMo word representation
and character representation for each word, 
and feed them into the \textbf{Bidirectional LSTM layer}~\cite{DBLP:conf/naacl/LampleBSKD16}. 
The outputs of both directions are concatenated together and form the word
representations $[\mathbf{h}_1, \mathbf{h}_2, \ldots, \mathbf{h}_n]$.

\subsubsection{Context-Sensitive span representation}

Given a span $c = \langle i, j\rangle$,
it separates the sentence into three parts and we can get their corresponding
hidden representations from the BiLSTM layer as:
\begin{inparaenum}[(1)]
\item the left context of the span
  $\mathbf{H}_{\text{lc}} = [\mathbf{h}_1,\ldots, \mathbf{h}_{i-1}]$; 
\item the inner content of the span $\mathbf{H}_{\text{c}} = [\mathbf{h}_i,\ldots,\mathbf{h}_{j}]$;
  and
\item the right context of the span
  $\mathbf{H}_{\text{rc}} = [\mathbf{h}_{j+1},\ldots,\mathbf{h}_{n}]$.
\end{inparaenum}

To obtain a fixed representation for any of $\mathbf{H}_{z}$
($z = \{\text{lc}, \text{c}, \text{rc}\}$), we adopt the self-attention
mechanism~\cite{DBLP:conf/iclr/LinFSYXZB17}. Specifically, we calculate the
attention vector, $\mathbf{a}_{z}$ over the variable-length input $\mathbf{H}_{z}$, and
then obtain the weighted hidden state $\mathbf{m}_{z}$ as $\mathbf{m}_{z} = \mathbf{H}_{z}\mathbf{a}^\top_{z}$, 
where $\mathbf{a}_{z} = \textsf{Softmax}(\mathbf{w}_{z}^\top \tanh(\mathbf{W}\mathbf{H}_{z}))$.
Finally, we obtain the final representation $\mathbf{m}$ by concatenating $\mathbf{m}_{\text{lc}}$,
$\mathbf{m}_{\text{c}}$, and $\mathbf{m}_{\text{rc}}$ together. 

\subsubsection{Loss function}

Given the final span representation $\mathbf{m}$, we multiply it with the type
embedding $\mathbf{l}$ of $l \in L$, and obtain the probability distribution over all types
via the $\textsf{Softmax}$ function:
\begin{equation}
  p(l \mid \mathbf{m}) = \frac{\exp(\mathbf{l}^\top\mathbf{m})}{\sum_{\tilde{\mathbf{l}} \in L}\exp(\tilde{\mathbf{l}}^\top \mathbf{m})},
  \label{eq:pre}
\end{equation}
The type embeddings $\mathbf{l}$ will be part of the parameters of the model. 

As one span can be annotated with multiple pre-defined types $l$ in $L$ and
the corresponding weight $w$, we modify the cross-entropy loss to take these into
consideration. The loss function $J$ is defined as
$ J = -\sum_{c\in C} \left( \sum_{l \in L} w_l \cdot \log(p(l \mid \mathbf{m})) \right)$,
where $C$ indicates all the candidate spans, $w_l$ indicates the annotation weight of type $l$ mentioned in Section~\ref{sec:overview}.

\subsection{Inference}
\label{sec:inference}
Our neural network based model provides probability estimation for \emph{all}
the candidate spans and \emph{all} the types, we need to perform
two tasks in the inference process:
\begin{inparaenum}[(1)]
\item find a non-overlapping span partition of the sentence, and 
\item assign a type to each span.
\end{inparaenum}
The second step is straightforward, as we just assign the type with the largest
probability predicted by the model. Hence, we mainly focus on the first step
below.

We model this problem as an optimization problem, where the goal is to find a
non-overlapping span partition such that the joint probability of each span
being \texttt{None} type is \emph{minimized}. The intuition is that we want to
encourage the algorithm to find more entity mentions of types \emph{other than}
\texttt{None}, hence increasing the recall of the method. 
More specifically, our objective function is:
\begin{align*}
  \argmin\limits_{V\in\mathcal{V}} \sum_{c\in V}\log(p(\text{\texttt{None}} \mid c)),
\end{align*}
where each $V$ is a partitioning of the sentence into a set of disjoint and
complete spans $[c_1, c_2, \ldots]$, and $\mathcal{V}$ is all the possible
partitionings.

\begin{algorithm}[htbp]
\SetKwFor{ForEach}{for each}{do}{end for}%
\SetKwInOut{Input}{Input}
\SetKwInOut{Output}{Output}

\small%
\Input{Candidate set $U$, sentence length $N$} 
\Output{Sentence $S$ with labeled spans}
$r[0]\gets0$;

\For{$j=1$ to $N$}{
    $r[j] \gets \min_{1\leq i \leq j}(r[i-1]+\log(p(\text{\texttt{None}}|\langle i,j\rangle))$;

}
$V \gets \mathsf{construct\_partitioning}(r)$;

$S \gets \mathsf{label\_spans}(V)$;

\Return{$S$}
  \caption{DP Based Inference}
  \label{alg:path-search}
\end{algorithm}

A naive optimization algorithm is to enumerate all possible partitionings and
compute the objective function. It is extremely costly as there are $O(2^{N})$
such partitionings ($N$ is the length of the sentence). In contrast, we propose
a dynamic programming algorithm, whose time complexity is only $O(NM)$~($M$ is the maximal 
number of words in a span), to
find the optimal partitioning.
As shown in Algorithm~\ref{alg:path-search}, 
where candidate set $U$ stores all spans and the predicted probability of the types, 
we use $r[j]$ to store the minimum
log probability of the
first $j$ tokens in the sentence. We sequentially update $r[j]$ where $j$
increases from $1$ to $N$. Note that $p(\text{\texttt{None}} \mid \langle{i, j}\rangle)$
can be computed in our model by computing the corresponding $\mathbf{m}$ for the span
$\langle{i, j}\rangle$ and Equation~\eqref{eq:pre}.

Once we obtained $r[N]$, we backtrack and reconstruct the optimal partitioning
(i.e., $\mathsf{construct\_partitions}$), and assign each span the type
with the highest probability (i.e., $\mathsf{label\_spans}$).

\section{Experiments}
In this section, we evaluate our method and compare it with other
supervised and distantly supervised methods on three benchmark datasets. In addition, 
we investigate the effectiveness of the designed method with a detailed analysis. 
\label{sec:experiments}
\subsection{Experiment Setting}

\subsubsection*{Datasets}

An overview of the datasets and dictionaries is shown in Table~\ref{tbl:dataset}.

\begin{table}[htp]
\centering
\caption{Dataset overview}
\resizebox{.95\columnwidth}{!}{
    \begin{tabular}{l c c c}
    \toprule
    Dataset & BC5CDR & NCBI-Disease & LaptopReview\\
    \midrule
    \midrule
    Domain & Biomedical & Biomedical & Technical Review\\
    \midrule
    Entity Types & \texttt{Disease}, \texttt{Chemical} & \texttt{Disease} & \texttt{AspectTerm} \\
    \midrule
    Dictionary & MeSH + CTD & MeSH + CTD & Computer Terms\\
    \bottomrule
    \end{tabular}
}
\label{tbl:dataset}
\end{table}

\begin{table*}[th]
\small
\centering
\caption{NER performance comparison. 
The \textbf{bold-faced} scores represent the best results among distantly supervised methods. }
\resizebox{2\columnwidth}{!}{
    \begin{tabular}{l lrrr rrr rrr}
    \toprule
    \multirow{2}{*}{Method} & Human Effort & \multicolumn{3}{c}{BC5CDR}  & \multicolumn{3}{c}{NCBI-Disease} & \multicolumn{3}{c}{LaptopReview}\\
    \cmidrule{3-11}
    & other than Dictionary & \multicolumn{1}{c}{Pre} & \multicolumn{1}{c}{Rec} & \multicolumn{1}{c}{$F_1$} & \multicolumn{1}{c}{Pre} & \multicolumn{1}{c}{Rec} & \multicolumn{1}{c}{$F_1$} & \multicolumn{1}{c}{Pre} & \multicolumn{1}{c}{Rec} & \multicolumn{1}{c}{$F_1$}\\
    \midrule
    \midrule
    Supervised model & Gold Annotations & 88.17 & 88.39 & 88.28 & 85.34 & 90.94 & 88.05 & 85.14 & 80.58 & 82.80\\
    \midrule
    \midrule
    \multirow{2}{*}{SwellShark} & Regex Design + Case Tuning  & 86.11 & 82.39 & 84.21 & 81.6 & 80.1 & 80.8 & - & - & - \\
    \cmidrule{2-11}
    & Regex Design  & 84.98 & 83.49 & 84.23 & 64.7 & 69.7 & 67.1 & - & - & - \\
    \midrule
    \midrule
    Dictionary Match & \multirow[c]{5}{*}{None} & \textbf{93.93} & 58.35 & 71.98 & \textbf{90.59} & 56.15 & 69.32 & \textbf{90.68} & 44.65 & 59.84\\
    \cmidrule{3-11}
    AutoNER* &  & 83.08 & 82.16 & 82.70 & 76.98 & 74.65 & 75.78 & 68.72 & 59.39 & 63.70\\
    \cmidrule{3-11}
    \our &  & \bccdrthep & \textbf{\bccdrther} & \textbf{\bccdrthef} & \ncbithep & \textbf{\ncbither} & \textbf{\ncbithef} & \laptopthep & \textbf{\laptopther} & \textbf{\laptopthef}\\
    \bottomrule
    \end{tabular}
}
\label{tbl:bioner}
\end{table*}

\texttt{\bccdr} dataset~\cite{DBLP:journals/biodb/LiSJSWLDMWL16} consists of 1,500 PubMed articles,
which has been separated into training set~(500), development set~(500), and test set~(500). 
The dataset contains 15,935 \texttt{Chemical} and 12,852 \texttt{Disease} mentions. 

\texttt{\ncbi} dataset~\cite{DBLP:journals/jbi/DoganLL14} consists of 793 PubMed abstracts, 
which has been separated into training set~(593), development set~(100), and test set~(100). 
The dataset contains 6,881 \texttt{Disease} mentions. 

\texttt{\laptop} dataset~\cite{DBLP:conf/semeval/PontikiGPPAM14} refers to
Subtask 1 for laptop aspect term (e.g., \textit{disk drive}) recognition. It
consists of 3,845 review sentences, which contains 3,012 \texttt{AspectTerm}
mentions.  
Following previous work~\cite{DBLP:conf/wassa/GiannakopoulosM17}, we separated
them into training set~(2445), development set~(609) and test set~(800).

\subsubsection*{Dictionary and High-Quality Phrase}
To fairly compare with previous methods, we construct and process dictionaries and high
quality phases in the same way as in~\cite{DBLP:conf/emnlp/ShangLGRR018}.
For the \bccdr and \ncbi datasets, we combine the MeSH
database~\footnote{\url{https://www.nlm.nih.gov/mesh/download_mesh.html}} and  
the CTD Chemical and Disease
vocabularies~\footnote{\url{http://ctdbase.org/downloads/}} as the dictionary. 
The phrases are mined from titles and abstracts of PubMed papers using 
the phrase mining method proposed in~\cite{DBLP:journals/tkde/ShangLJRVH18}.

For the \laptop dataset, we crawl dictionary terms from the public website~\footnote{\url{https://www.computerhope.com/jargon.htm}}. 
The phrases are mined from the Amazon laptop review dataset\footnote{\url{http://times.cs.uiuc.edu/~wang296/Data/}} 
using the same phrase mining method.

As suggested in~\cite{DBLP:conf/emnlp/ShangLGRR018}, we apply tailored
dictionaries to reduce false positive matching.

\subsubsection*{Headword}
We use rule based method proposed in~\cite{DBLP:conf/acl/ZhouSZZ05} to extract the headwords of phrases.
The headword of a phrase is generally the last word of the phrase. 
If there is a preposition in the phrase, the headword is the last word before the preposition. 
For example, the headword of the phrase \textit{cancer of the liver} is \textit{cancer}.

\subsubsection*{Metric}
Following the standard setting, we evaluate the methods using micro-averaged $F_1$ score 
and also report the precision~(Pre) and recall~(Rec) in percentage. 
All the reported scores are averaged over five different runs.

\subsection{Implementation}
We use the same word embeddings for the neural model and the cosine similarity mentioned in Section~\ref{sec:overview}.
For the \bccdr and \ncbi datasets, we use pre-trained 200-dimensional word embeddings~\cite{moen2013distributional} and ELMo\footnote{\url{https://allennlp.org/elmo}} trained on a PubMed corpus.
For the \laptop dataset, we use pre-trained 100-dimensional GloVe word embeddings~\cite{DBLP:conf/emnlp/PenningtonSM14} and ELMo trained on a news corpus.
We select hyper-parameters that achieve the best performance on the development sets.
Details of hyper-parameters of dictionary extension, neural model, and training process are listed in the \emph{appendix}.
\subsection{Compared Methods}
\label{sec:compare}
We compare~\our with other methods from three different categories: supervised
model, distantly
supervised method with human effort, and distantly supervised method without human
effort. 

\noindent{\textbf{Supervised model.}} We use supervised model to demonstrate the
competitiveness of~\our. 
We use the state-of-the-art model architecture proposed
in~\cite{DBLP:conf/naacl/PetersNIGCLZ18} with ELMo trained in the corresponding
domains.

\noindent{\textbf{SwellShark}}~\cite{DBLP:journals/corr/Fries0RR17} is a
distantly supervised method designed for the biomedical domain, especially on
the \bccdr and \ncbi datasets. It requires human efforts to customize regular
expression rules and hand-tune candidates. 

\noindent{\textbf{AutoNER}}~\cite{DBLP:conf/emnlp/ShangLGRR018} is the previous
state-of-the-art distantly supervised method without human efforts on all the 
datasets. Similar to our method, it only requires 
domain specific dictionaries and a set of high-quality phrases. 
To make a fair comparison with \our, we have re-implemented
AutoNER to add ELMo as part of input representation, which brings the performance improvement. 

AutoNER is originally trained with all the raw texts from the
dataset~(i.e., the training corpus is the union of the training set, development set,
 and test set). 
To make a fair comparison with all other methods, 
we have re-trained AutoNER with raw texts only from the training set. This
brings a lower performance than those in the original paper.

Therefore, we use \textbf{AutoNER*} for AutoNER with ELMo and trained on the training
set only, and report the evaluation results of AutoNER* in our paper.
The readers are referred to~\cite{DBLP:conf/emnlp/ShangLGRR018} for the original
performance of AutoNER. 
Nevertheless, we have tried to train AutoNER with ELMo and using all the raw
texts, and the performance is still lower than \our.

We also include purely dictionary match method. We directly apply the original dictionaries 
to the test sets and get entity mentions and the types by exact string matching.

\subsection{Overall Performance}

Table~\ref{tbl:bioner} shows the performance of
different methods on all the datasets. 
\our achieves the best recall and $F_1$ among all the distantly
supervised methods.
It surpasses SwellShark by around 2\% $F_1$ score on the \bccdr and \ncbi
datasets, though SwellShark is specially designed for the biomedical domain
and needs human efforts.

Purely dictionary matching method achieves the highest precision among all the
methods, which is not surprising as dictionaries always contain accurate entity
mentions. However, it suffers from low recall which is due to the low coverage
of the dictionary and the strict matching rule.

On the contrary, the success of \our owes to the ability of generalization on
the out-of-dictionary entity mentions without losing prediction accuracy. 
As the result, \our achieves significant improvement over the
previous state-of-the-art method (i.e., AutoNER*) on both precision and recall.
The $F_1$ scores are improved by 3.47\%, 7.01\%, and 3.76\% on the three
datasets, respectively.

Detailed analysis is shown in the following sections.

\subsection{Effectiveness of Dictionary Extension}
\label{sec:effect_dict_extension}
\begin{table*}[htp]
  \small
  \centering
  \caption{Effectiveness of each component.
    We do not apply annotation weights on AutoNER* as it does not support weighted annotations.
    }
    \begin{tabular}{lrrrrrrrrr}
    \toprule
    Component & \multicolumn{3}{c}{BC5CDR} & \multicolumn{3}{c}{NCBI-Disease} & \multicolumn{3}{c}{LaptopReview}  \\
    \cmidrule{2-10}
    \multicolumn{1}{c}{} & \multicolumn{1}{c}{Pre} & \multicolumn{1}{c}{Rec} & \multicolumn{1}{c}{$F_1$} & \multicolumn{1}{c}{Pre} & \multicolumn{1}{c}{Rec} & \multicolumn{1}{c}{$F_1$} & \multicolumn{1}{c}{Pre} & \multicolumn{1}{c}{Rec} & \multicolumn{1}{c}{$F_1$} \\
    \midrule
     Neural Model&\bccdrtp&\bccdrtr&\bccdrtf&\ncbitp&\ncbitr&\ncbitf&\laptoptp&\laptoptr&\laptoptf\\

     Neural Model+Dictionary Extension &\bccdrthp&\bccdrthr&\bccdrthf&\ncbithp&\ncbithr&\ncbithf&\laptopthp&\laptopthr&\laptopthf\\
     Neural Model+ELMo&\bccdrtep&\bccdrter&\bccdrtef&\ncbitep&\ncbiter&\ncbitef&\laptoptep&\laptopter&\laptoptef\\
    \midrule
     Neural Model+Dictionary Extension+ELMo&\textbf{\bccdrthep}&\textbf{\bccdrther}&\textbf{\bccdrthef}&\textbf{\ncbithep}&\textbf{\ncbither}&\textbf{\ncbithef}&\textbf{\laptopthep}&\textbf{\laptopther}&\textbf{\laptopthef}\\
    \midrule
    AutoNER*+Dictionary Extension & 81.53 & 83.03 & 82.28& 80.12 & 83.02& 81.54 &69.61& 59.45 & 64.11\\
    \bottomrule
    \end{tabular}    
    \label{tbl:abl}
  \end{table*}
We explore the effectiveness of dictionary extension proposed in Section~\ref{sec:overview}.
We use precision and recall to evaluate the accuracy and the coverage of
annotations on the training set. 
As shown in Table~\ref{tbl:dict}, using the extended dictionary is able to boost
the recall~(coverage) by a large margin (e.g., on the~\bccdr and \ncbi datasets,
the recall increases by more than 20\%), while slightly reducing the precision
(which is inevitable due to the additional noise introduced by the extended
dictionary terms).
\begin{table}[htp]
\small
\centering  
\caption{Annotation quality on the training set.}
\resizebox{.95\columnwidth}{!}{
    \begin{tabular}{lrrrrrr}
    \toprule
    \multicolumn{1}{l}{\multirow{2}{*}{}} & \multicolumn{2}{c}{BC5CDR} & \multicolumn{2}{c}{NCBI-Disease} & \multicolumn{2}{c}{LaptopReview}  \\
    \cmidrule{2-7}
    \multicolumn{1}{c}{} & \multicolumn{1}{c}{Pre} & \multicolumn{1}{c}{Rec} & \multicolumn{1}{c}{Pre} & \multicolumn{1}{c}{Rec} & \multicolumn{1}{c}{Pre} & \multicolumn{1}{c}{Rec} \\
    \midrule
    Dictionary &94.92&72.23&93.88&60.56&89.01&51.65\\
    \midrule
    Extended Dictionary &91.89&84.48&92.61&83.01&87.73&54.51\\
    \bottomrule
    \end{tabular}
}
    \label{tbl:dict}
\end{table}

\begin{table}[htp]
  \small
  \centering
  \caption{Effectiveness of weighted annotations. 
    }
\resizebox{.95\columnwidth}{!}{
    \begin{tabular}{llrrrrrrrrr}
    \toprule
    \multicolumn{2}{c}{\multirow{2}{*}{Method}} & \multicolumn{3}{c}{BC5CDR} & \multicolumn{3}{c}{NCBI-Disease} & \multicolumn{3}{c}{LaptopReview}  \\
    \cmidrule{3-11}
    \multicolumn{2}{c}{} & \multicolumn{1}{c}{Pre} & \multicolumn{1}{c}{Rec} & \multicolumn{1}{c}{$F_1$} & \multicolumn{1}{c}{Pre} & \multicolumn{1}{c}{Rec} & \multicolumn{1}{c}{$F_1$} & \multicolumn{1}{c}{Pre} & \multicolumn{1}{c}{Rec} & \multicolumn{1}{c}{$F_1$} \\
    \midrule
    \multirow{2}{*}{\our}&with weight &\textbf{\bccdrthep}&\bccdrther&\textbf{\bccdrthef}&\textbf{\ncbithep}&\ncbither&\textbf{\ncbithef}&\textbf{\laptopthep}&\textbf{\laptopther}&\textbf{\laptopthef}\\
    \cmidrule{2-11}
     &without weight &\bccdrthehp&\textbf{\bccdrthehr}&\bccdrthehf&\ncbithehp&\textbf{\ncbithehr}&\ncbithehf&\textbf{\laptopthehp}&\textbf{\laptopthehr}&\textbf{\laptopthehf}\\
    \bottomrule
    \end{tabular}
}

\label{tbl:func}
\end{table}

The increasing of recall on the \laptop dataset is not as significant as those
on the other two datasets. This is due to the impact of the similarity
threshold mentioned in Section~\ref{sec:overview}.
On the \laptop dataset, we use a higher similarity threshold to
avoid introducing too much noise to the extended dictionary, 
which brings less
improvement to the coverage of the extended dictionary.

We also investigate the effectiveness of the weight function~(i.e. Equation~\eqref{func:weight}) in the loss function.
We assign all annotations with weight 1 to eliminate the effect of the weight
function (noted as \our without weight), and show the results in Table~\ref{tbl:func}.
It can be seen that the weight function helps \our achieve better
precision~(e.g. 0.55\% and 1.4\% improvement on the \bccdr and \ncbi datasets, respectively)
with slightly decrease of recall. 

The reason why we have not observed the improvement on the \laptop dataset is
that we set a higher similarity threshold.
We observe the same trend as the other two datasets when we set a lower
similarity threshold.

\subsection{Effectiveness of Model}
\label{sec:ablation}

To demonstrate the effectiveness of each component in the neural model, we 
conduct experiments by disabling different components in the model on all the datasets.
From the results in Table~\ref{tbl:abl}, we observe that each component
improves the model from different aspects.

It is worth mentioning that even without using ELMo or the extended dictionaries, the neural model still
shows competitive performance compared with AutoNER* which extracts mention
boundaries and then predicts the types. 
This reveals that the span-level modeling and the inference algorithm is more suitable for
the NER task compared with the pipeline design.

Dictionary extension improves the coverage of the pseudo annotations, hence boosts the
coverage of predictions~(i.e. recall).
In addition, dictionary extension will not bring in many false-positive
predictions, hence will not affect precision.
For example, on the~\ncbi dataset, dictionary extension improves the precision and
recall by 0.58\% and 5.25\%, respectively. 
\begin{table}[htp]
\small
\centering
\caption{Comparison of $F_1$ scores on in-dictionary (ID) and
    out-of-dictionary (OOD) entity mentions.}
\resizebox{.95\columnwidth}{!}{
    \begin{tabular}{llllllll}
    \toprule
    \multicolumn{2}{c}{}& \multicolumn{2}{c}{BC5CDR} & \multicolumn{2}{c}{NCBI-Disease} & \multicolumn{2}{c}{LaptopReview} \\
    \cmidrule{3-8}
    & & \multicolumn{1}{c}{ID} & \multicolumn{1}{c}{OOD} & \multicolumn{1}{c}{ID} & \multicolumn{1}{c}{OOD} & \multicolumn{1}{c}{ID} & \multicolumn{1}{c}{OOD} \\
    \midrule
    \multicolumn{2}{c}{the number of entity mentions} &5734 &3991 &539 &416 &292 &362 \\
    \midrule
    \midrule
    \multirow{2}{*}{$F_1$} &\our &\textbf{93.53}&\textbf{76.05}
        &\textbf{91.96}&\textbf{70.32}
        &\textbf{92.86}&\textbf{45.51}\\
    \cmidrule{2-8}
    &AutoNER* &92.80 &66.41 
            &91.88&54.05
            &89.23&37.64\\
\bottomrule
    \end{tabular}
  }  
\label{tbl:oov}
\end{table}
On the other hand, ELMo focuses on the accuracy of predictions. For example,
the precision increases from \bccdrtp\% to \bccdrtep\% on
the~\bccdr dataset.
ELMo also encourages the model to extract more entity mentions.

With both components,
\our achieves the best overall performance. It seems that the improvement from
dictionary extension and ELMo are additive, 
which implies that dictionary extension and ELMo improve the model from different aspects.

In order to show the effectiveness of dictionary extension to other models, as
well as to show the effectiveness of our proposed framework, we apply the
extended dictionaries in AutoNER*. 
We observe that the performance of AutoNER* improves with the extended
dictionaries, however, it is still much lower than our proposed method \our{}.

\subsection{Generalization on OOD Entity Mentions}
\label{sec:oov}

We perform analysis on the test set for 
the out-of-dictionary entity mentions to investigate the generalization of~\our. 
Specifically, we partition entity mentions into two subsets: in-dictionary~(ID) entity mentions and
out-of-dictionary~(OOD) entity mentions. An entity mention is considered as an ID mention
if it is fully or partially appearing in the dictionary, and an OOD mention otherwise.

Table~\ref{tbl:oov} shows the performance of \our and AutoNER* on the OOD and ID
entity mentions. 
\our surpasses AutoNER* on the ID entity mentions, and it also shows significant
improvement on the OOD entity mentions over all the datasets. 
For example, on the \ncbi dataset, it boosts the $F_1$ score by at least 16\% on
the OOD entity mentions. 
This explains why \our has a higher overall performance and also demonstrates
the stronger generalization ability of \our.

\section{Conclusion}
\label{sec:conclusion}
In this paper, we presented a new method to tackle NER tasks in specific domains using distant supervision.
Our method exploits several ideas including headword-based dictionary extension,
span-level neural model, and dynamic programming inference. Experiments show 
\our significantly outperforms the previous state-of-the-art methods.

\paragraph*{Acknowledgement}%
This work is supported by ARC DPs 170103710 and 180103411, D2DCRC DC25002 and
DC25003, NSFC under grant No.\ 61872446 and PNSF of Hunan under grant No.\
2019JJ20024. The Titan V used for this research was donated by the NVIDIA
Corporation.

\bibliographystyle{aaai}
\bibliography{AAAI-LiuS.arxiv}

\section*{A Appendix}
To help readers reproduce the results, we represent the details for implementing our proposed method.
In this section, we list details of hyper-parameter for dictionary extension, neural model construction, and training.

\subsection*{A.1 Dictionary Extension Hyper-parameters}
Table~\ref{tbl:dictionaryextention} lists the hyper-parameters for dictionary extension.
In the biomedical domain, we set headword threshold $\tau_{\text{1}}$ to 5 and similarity threshold $\tau_{\text{2}}$ to 0.4.
In the technical review domain, we set headword threshold $\tau_{\text{1}}$ to 3 and similarity threshold $\tau_{\text{2}}$ to 0.9.
This controls the possible noise introduced by the extended dictionary terms.
For the weight function, we set $\theta_1$ to 1.0 and $\theta_2$ to -0.5 in both domains.
\begin{table}[htp]
\centering
\small
\caption{Dictionary Extension Hyper-parameters}
\resizebox{.95\columnwidth}{!}{
\begin{tabular}{c|cc}
\toprule
 Domain & Biomedical & Technical Review \\
\midrule
headword frequency threshold $\tau_{\text{1}}$ &5&3\\
\midrule
similarity threshold $\tau_{\text{2}}$ &0.4&0.9\\
\midrule
label weight $\theta_1$&\multicolumn{2}{c}{1.0}\\
\midrule
label weight $\theta_2$&\multicolumn{2}{c}{-0.5}\\
\bottomrule
\end{tabular}
}

\label{tbl:dictionaryextention}
\end{table}

\subsection*{A.2 Neural Model Hyper-parameters} 
Table~\ref{tbl:modelparameters} lists the hyper-parameters for the neural model.
Character embeddings are initialized with uniform samples from $[-\sqrt{\frac{3}{d}},
+\sqrt{\frac{3}{d}}]$, with $d$ is set to 16. 
We apply CNNs with kernel sizes 2,3,4 to fully explore the morphological information in the biomedical domain.
In the technical review domain, we use CNNs with kernel size 3.
For the \bccdr and \ncbi datasets, we use pre-trained 200-dimensional word embeddings~\cite{moen2013distributional} and ELMo\footnote{\url{https://allennlp.org/elmo}} trained on a PubMed corpus.
For the \laptop dataset, we use pre-trained 100-dimensional GloVe word embeddings~\cite{DBLP:conf/emnlp/PenningtonSM14} and ELMo trained on a news corpus.
We use a single-layer Bidirectional LSTM~(BiLSTM)~\cite{DBLP:conf/naacl/LampleBSKD16} with
the hidden state dimension of the BiLSTM set to 256.
\begin{table}[htp]
\centering
\small
\caption{Neural Model Hyper-parameters}
\resizebox{.95\columnwidth}{!}{
\begin{tabular}{c|c|cc}
\toprule
 \multicolumn{2}{c|}{Domain} & Biomedical & Technical Review  \\
\midrule
Char-level Embedding            &  dimension & \multicolumn{2}{c}{16} \\
\midrule
\multirow{4}{*}{CNN} & kernel size & 2,3,4 & 3 \\
                      & padding & 1,2,3 & 2 \\
                       & stride & 1,1,1 & 1 \\
                       & channel & 128,128,128 & 128 \\
\midrule
Word-level Embedding            & dimension & 200 & 100\\
\midrule
ELMo & dimension & \multicolumn{2}{c}{1024}\\
\midrule
\multirow{2}{*}{BiLSTM} & hidden size & \multicolumn{2}{c}{256}\\
                       & layer & \multicolumn{2}{c}{1}\\
\midrule
\multirow{3}{*}{Dropout Rate} & embedding & \multicolumn{2}{c}{0.33}\\
& BiLSTM input & \multicolumn{2}{c}{0.33}\\
& BiLSTM output & \multicolumn{2}{c}{0.5}\\
\midrule
Attention & dimension & \multicolumn{2}{c}{256} \\
 \bottomrule
\end{tabular}
}

\label{tbl:modelparameters}
\end{table}

\subsection*{A.3 Training Hyper-parameters} 
Table~\ref{tbl:trainparameters} lists the parameter used during training.
We process at most 1000 tokens in each batch and optimize the model parameters using Adam~\cite{DBLP:journals/corr/KingmaB14} optimizer with $\beta_1 = 0.9$ and $\beta_2 = 0.999$.
The initial learning rate is set to 0.001.
We use gradient clipping of 5.0 for better stability.
The maximal number of words in a span~(i.e., $M$) is set to 5 for all the datasets, which covers more than 99\% entity mentions based on statistic information from the development set.
To further reduce the number for non-entity spans during training, 
we restrict the ratio of negative spans~(i.e. annotated with \texttt{None} type) and positive spans~(i.e. annotated with types other than \texttt{None}) to be $5:1$ by
uniformly sampling negative spans.
During prediction, we generate all the possible spans with no more than $M$ words for each sentence.
\begin{table}[htp]
\centering
\small
\caption{Training Hyper-parameters}
\resizebox{.95\columnwidth}{!}{
\begin{tabular}{c|cc}
\toprule
 Domain & Biomedical & Technical Review \\
\midrule
Number of tokens per batch&\multicolumn{2}{c}{1000} \\
\midrule
Number of epoch &\multicolumn{2}{c}{50} \\
\midrule
Optimizer& \multicolumn{2}{c}{Adam}\\
learning rate & \multicolumn{2}{c}{0.001}\\
Adam $\beta_1$ & \multicolumn{2}{c}{0.9}\\
Adam $\beta_2$ & \multicolumn{2}{c}{0.999}\\
\midrule
gradient clip& \multicolumn{2}{c}{5}\\
\midrule
maximal number of words per span $M$&\multicolumn{2}{c}{5}\\
\midrule
non-entity span rate&\multicolumn{2}{c}{5:1}\\
\bottomrule
\end{tabular}
}

\label{tbl:trainparameters}
\end{table}

\end{document}